\documentclass[runningheads]{llncs}

\usepackage{amsmath}
\usepackage{caption,subcaption}
\usepackage{pgffor}
\usepackage{csquotes}
\usepackage[T1]{fontenc}
\usepackage[utf8]{inputenc}

\usepackage{multirow}
\usepackage{microtype}
\usepackage[english]{babel}

\usepackage[table]{xcolor}
\usepackage{hyperref} 

\usepackage{booktabs}
\usepackage{tabularx}
\usepackage{ragged2e}
\usepackage{longtable} 
\usepackage{afterpage}
\usepackage{rotating}
\usepackage{enumitem}
\usepackage{graphicx}
\usepackage[most]{tcolorbox}
\usepackage{calc}

\usepackage[export]{adjustbox}

\newcommand{\character}[1]{{‹#1›}}

\newlength\myheight
\newlength\mydepth
\settototalheight\myheight{Xygp}
\newcommand*\inlinegraphics[1]{%
  \settototalheight\myheight{Xygp}%
  \settodepth\mydepth{Xygp}%
    {\centering\includegraphics[height=\myheight]
    {#1}}%
}

\usepackage{tikz}
\def\checkmark{\tikz\fill[scale=0.4](0,.35) -- (.25,0) -- (1,.7) -- (.25,.15) -- cycle;}

\newcommand{\longs}{
  \tikz[baseline={(0,0)}]
  \node[anchor=base, inner sep=0pt, outer sep=0pt] {
    \begin{tikzpicture}[scale=0.1, line cap=round, line join=round]
      \draw[thick] 
      (0,0) -- (0,2) 
      .. controls (0.1,2.5) and (0.8,2.3) .. (0.5,2.3);
    \end{tikzpicture}
  };
}

\begin{document}

\title{An Interpretable Deep Learning Approach \\ for Morphological Script Type Analysis}
\titlerunning{An Interpretable Deep Learning Approach for Script Type Analysis}
%
\author{Malamatenia Vlachou-Efstathiou\inst{1,2}\orcidID{0000-0002-9397-356} \and 
Ioannis Siglidis\inst{2}\orcidID{0009-0002-2278-5825}
\and
Dominique Stutzmann\inst{1}\orcidID{0000-0003-3705-5825} \and
Mathieu Aubry\inst{2}\orcidID{0000-0002-3804-0193} }

\authorrunning{M. Vlachou-Efstathiou et al.}
%
\institute{Institut de Recherche et d'Histoire des Textes, Paris, Île-de-France, France \\
\email{\{malamatenia.vlachou, dominique.stutzmann\}@irht.cnrs.fr} \and
LIGM, Ecole des Ponts, Univ Gustave Eiffel, CNRS, Marne-la-Vallée, France \\
\email{\{mathieu.aubry, ioannis.siglidis\}@enpc.fr}
\url{https://learnable-handwriter.github.io/}}
\maketitle  

\begin{abstract} 

Defining script types and establishing classification criteria for medieval handwriting is a central aspect of palaeographical analysis. However, existing typologies often encounter methodological challenges, such as descriptive limitations and subjective criteria. We propose an interpretable deep learning-based approach to morphological script type analysis, which enables systematic and objective analysis and contributes to bridging the gap between qualitative observations and quantitative measurements. More precisely, we adapt a deep instance segmentation method to learn comparable character prototypes, representative of letter morphology, and provide qualitative and quantitative tools for their comparison and analysis. We demonstrate our approach by applying it to the \textit{Textualis Formata} script type and its two subtypes formalized by A. Derolez: Northern and Southern \textit{Textualis}.

\end{abstract}
%





\section{Introduction}

The concept of \textit{script type} is of central importance to palaeography, which studies handwritten documents in relation to their context of production such as date, origin, and scribal hands, to support historical discourse. Adapting M. Parkes, we define a \textit{script type} as \enquote{the model which the scribe has in their mind's eye when they write}~\cite{parkes1969script,stokes2011_describing_handwriting}, that is, if we restrict ourselves to characters, the set of prototypical forms of each character towards which they are working when they write. In order to discretize the \textit{continuum} of handwritten forms and establish script types and their classification criteria, the palaeographical method compares handwriting samples, describes, and analyzes the variations of letter forms. Palaeographers often resort to the idea of script types as \textit{ideal prototypes}~\cite{lineamenti_storia_scrittura_latina}, with the delineation of artificial alphabets, i.e., sets of abstracted letter forms. Several typologies have been proposed and refined over the years. Most recently A. Derolez proposed a taxonomy for Gothic book scripts based on letter morphology~\cite{derolez2003palaeography}, where some letters with distinctive visual elements serve as the basis for classification. 

In this paper, we introduce a methodology that leverages deep learning for the analysis of morphological script types. 
More precisely, we learn aligned character prototypes from documents and present different methods for qualitative and quantitative analysis. This enables us to confront different documents to existing typologies, potentially adding nuance or complementing them. Indeed, existing typologies present persistent methodological issues~\cite{derolez2003palaeography,smith2004derolez,stutzmann_nomenklatur_2005}, mainly the ambiguity arising from relying on a \enquote{global impression} to discern scripts, inconsistencies in nomenclature across scholarly traditions, and difficulties in describing minute morphological differences using natural language~\cite{gasparri1966remarques,stokes2011_describing_handwriting}. These challenges underscore the potential benefits of methods such as ours, which could enhance palaeographical analysis through a systematic and objective approach, and facilitate the integration of quantitative measures with qualitative observations.

This is in line with the position of pioneers like Léon Gilissen~\cite{gilissen1973expertise,gilissen1975rapportmodulaire} and others~\cite{poulle1974paleographie,ornato1975,sirat1981examen,tomiello1996dalla,davis1998towards,mcgillivray2005statistical,stutzmann_paleographie_2010,muzerelle_analyse_2011}, who experimented with statistical measurements and the modeling of \textit{measurable elements} of script. Such measurements of scripts pose significant challenges, such as defining a set of descriptors or discriminative handwriting features and ensuring comparable objects and magnitudes~\cite{stansbury2009computer}. Contrary to classification tasks such as writer and geographical attribution, which are formulated as discriminative learning problems, script type analysis cannot be reduced to a classification problem~\cite{stutzmann_systeme_2013,hassner2015computation,stutzmann_clustering_2016} and adequate modeling of variations is crucial~\cite{stutzmann2018variability}. 
Indeed simply matching external samples to pre-defined script types does not help better understanding and questioning the classification criteria.

We thus propose a method for evidence-based paleography focusing on interpretability rather than script classification. Our key idea is to remain close to classical morphological approaches for defining taxonomies and introduce tools to model and analyze letter shapes automatically. We build on the Learnable Typewriter approach~\cite{the-learnable-typewriter} and adapt it so that it can learn comparable character prototypes, which requires designing appropriate finetuning strategy and filtering. We then introduce visualizations and graphical tools, as well as an interpretable variability measure. To demonstrate how such tools can be leveraged for palaeographic analysis, we select a corpus in \textit{Textualis Formata} and present a case study on the morphological analysis of its two subtypes, Northern and Southern \textit{Textualis}.

\noindent \textbf{Contribution.} In summary, our main contributions are: 
\begin{itemize}[topsep=0pt, partopsep=0pt, itemsep=0pt, parsep=0pt]
    \item the adaptation of a deep instance segmentation method for palaeographical script type analysis
    \item a methodology for homologous comparison of characters, including visualization, graphical, and quantitative tools
    \item a case study demonstrating how these tools can complement the classic taxonomy of A. Derolez~\cite{derolez2003palaeography} for the analysis of Northern and Southern \textit{Textualis}.
\end{itemize}

\section{Related work}

We first give an overview of works that develop quantitative methods for palaeographic analysis. We then present \enquote{prototype-based} approaches to document analysis, which, although not specifically developed for paleographic analysis, are the basis of our approach.

\begin{figure}[t]
    \centering    \includegraphics[width=1\textwidth]{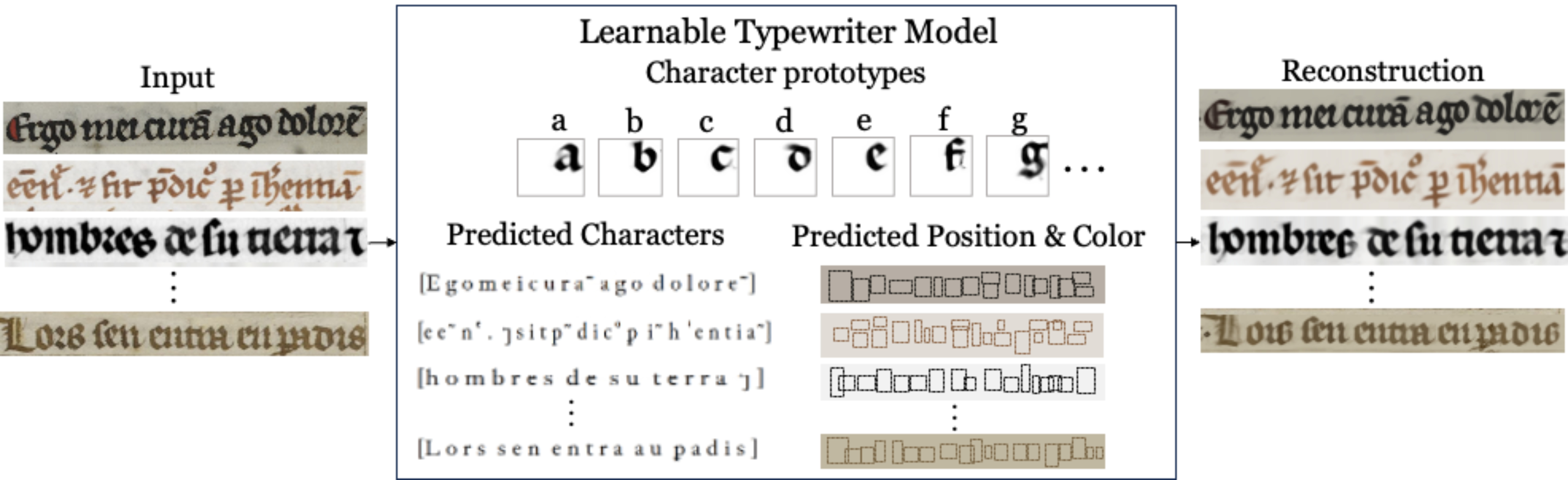}
    \caption{\textbf{The Learnable Typewriter Model} learns to reconstruct text lines using a set of learned character prototypes. We demonstrate how the character prototypes can be used for palaeographic analysis.}
    \label{fig:learnable-typewriter}
\end{figure}

\paragraph{Quantitative methods in palaeography.}

In the past two decades, many automatic methods have been developed for writer or script classification~\cite{sommerschield2023machine,nigam2023document}, using texture-based features~\cite{moalla2006image,schomaker2007using,wolf2011automatic,lebourgeois_caracterisation_2011,he2014delta,djeddi2014evaluation,hannad2016writer}, grapheme-based features~\cite{schomaker2004automatic,siddiqi2010text,djeddi2013codebook,he2015junction} and deep learning classification approaches~\cite{cloppet_new_2012,christlein2015offline,cloppet_icdar_2017,cloppet_icfhr2016_2016,tang2016text,stutzmann_icdar_2017,he2019deep}. Some papers, such as~\cite{kestemont2017artificial}, make a particular effort to build interpretable features or to visualize deep features responsible for the classification, but their interpretation remains limited.

Another branch of studies aims at producing interpretable outputs for palaeographic analysis of letter structure. The Information System for Graphological Identification~\cite{mamatsis2023novel}, extracts the average shape of specific characters via curve and contour detection, standardizing orientation and size, for automatic hand comparison and writer identification. The Graphem project~\cite{muzerelle_analyse_2011} focused specifically on script type features. \cite{eglin2011outils} explores visually interpretable stroke analysis, by extracting connected components to create a strokes code book, and then grouping the strokes through graph coloring for categorization of elementary stroke shapes. Closest to us, focusing on entire letter form variations for script type analysis is the System of Palaeographical Inspection~\cite{aiolli1999spi,ciula2005SPI}. It generates an average character prototype by computing the centroid of semi-automatically segmented occurrences. The prototypes are used both for hierarchical clustering of similar hands and classification of external samples.

However, the results of these approaches can hardly be compared with minute traditional palaeographic analysis and do not provide complete automation. Instead, our idea is to build on methods that directly learn prototypical characters from documents and use them for actual paleographic analysis.

\paragraph{Prototype-based approaches in document analysis.} Early methods for document analysis~\cite{hochberg1997automatic,kopec1997supervised,xu1999prototype,baird1999model,Ocular2013} use variants of character template matching for analyzing documents. While their main goal is often to perform optical character recognition (OCR), such methods typically also produce finer outputs, such as character segmentation, and learn a character template or prototype.
Similar approaches have thus been used for typographical analysis of early prints~\cite{ramel2013interactive,goyal2020probabilistic,kordon2023classification}.
This type of approach has recently been revisited with deep learning tools by the Learnable Typewriter approach~\cite{the-learnable-typewriter}. We build on this method and describe it in the next section.

\section{Approach}

\subsection{Learning comparable character prototypes}\label{methodology}
\label{sec:ltw}
\paragraph{The Learnable Typewriter.}
We build our approach on the Learnable Typewriter model~\cite{the-learnable-typewriter}, visualized in Figure~\ref{fig:learnable-typewriter}. This deep learning model learns to reconstruct text lines by compositing a set of \textit{character prototypes} on a simple background. Given as input the image of a line, it predicts the {color of the background}, the {characters} used in the line, and for each character, its {position} and {color}. The {character prototypes} are also learned by the model, and each instance of a character is reconstructed with the exact same prototype. The model can be trained, as in our experiments, using a set of text line images with their transcriptions. 

Each character prototype is a grayscale image and can be thought of as the average shape of all occurrences of a character in the training data, standardized for color, size, and position. Therefore, training a Learnable Typewriter model on a particular corpus, such as one corresponding to a specific script or handwriting style, will yield the average shape of each character without the need for manual selection of specific character samples, annotation of character positions, or binarization.

\paragraph{Finetuning character prototypes.}\label{training-finetuning} We propose to compare different documents and different {scripts} by comparing the character prototypes learned on various corpora. However, directly comparing prototypes learned by different models is not possible, because they are not aligned. Our solution is first to learn a \textit{reference model} using a reference corpus - in our case study, a set of documents in \textit{Textualis Formata} - then finetune the model to reconstruct selected documents, keeping all network parameters frozen except those that only impact the prototypes. Since the positioning, scaling, and coloring of the prototypes are shared, the prototypes will remain aligned and can be directly compared, such as by computing their difference. We define a single reference corpus to obtain reference prototypes, and then finetune them on multiple specific corpora, that may or may not be part of the reference corpus.

\paragraph{Character prototype filtering.}\label{denoising}

\begin{figure}[t]
    \centering
    \begin{subfigure}[t]{0.48\textwidth}
        \centering
        \includegraphics[width=\textwidth]{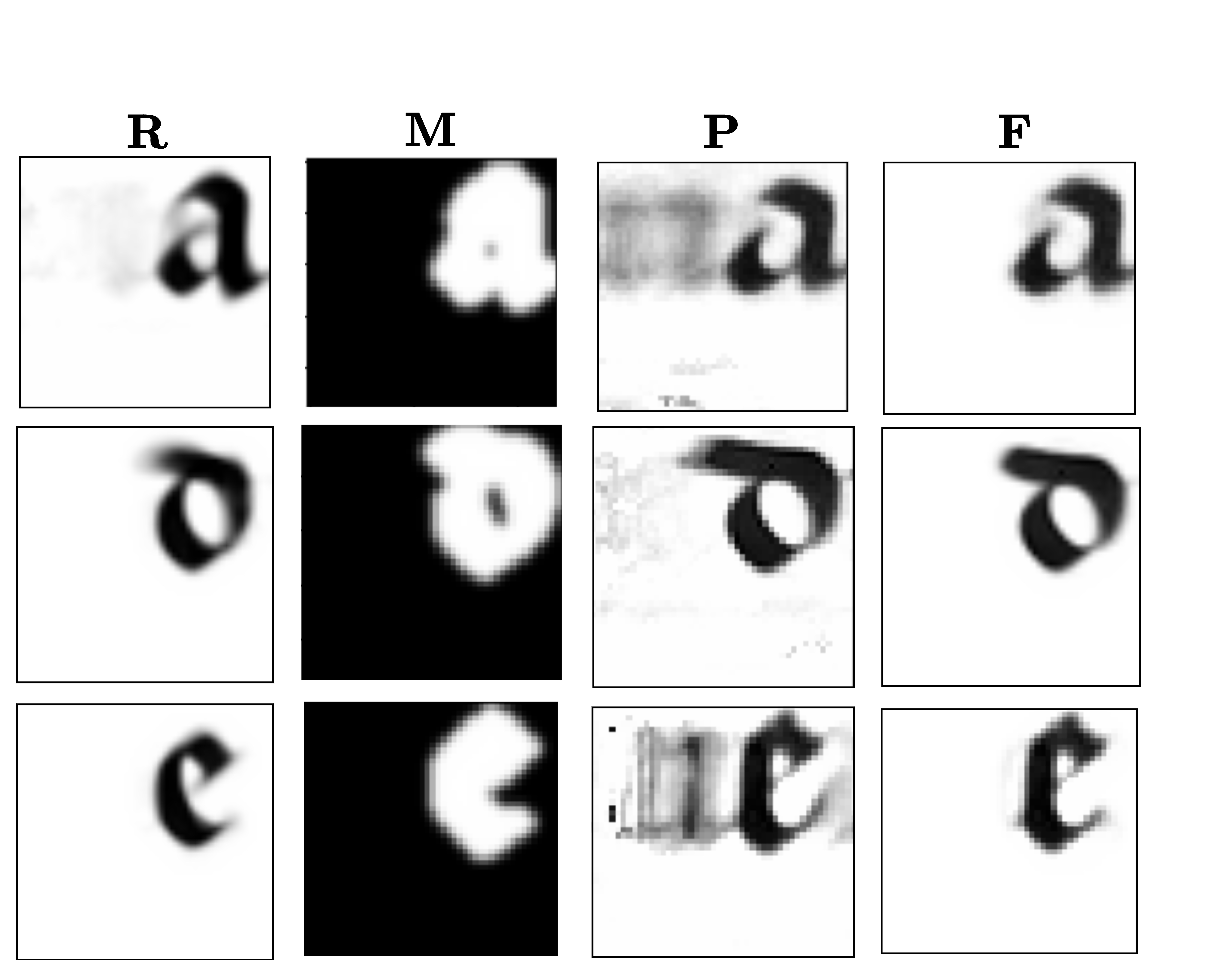}
        \caption{\textbf{Filtering strategy}}
        \label{fig:graph-denoising}
    \end{subfigure}
    \begin{subfigure}[t]{0.47\textwidth}
        \centering
        \includegraphics[width=\textwidth]{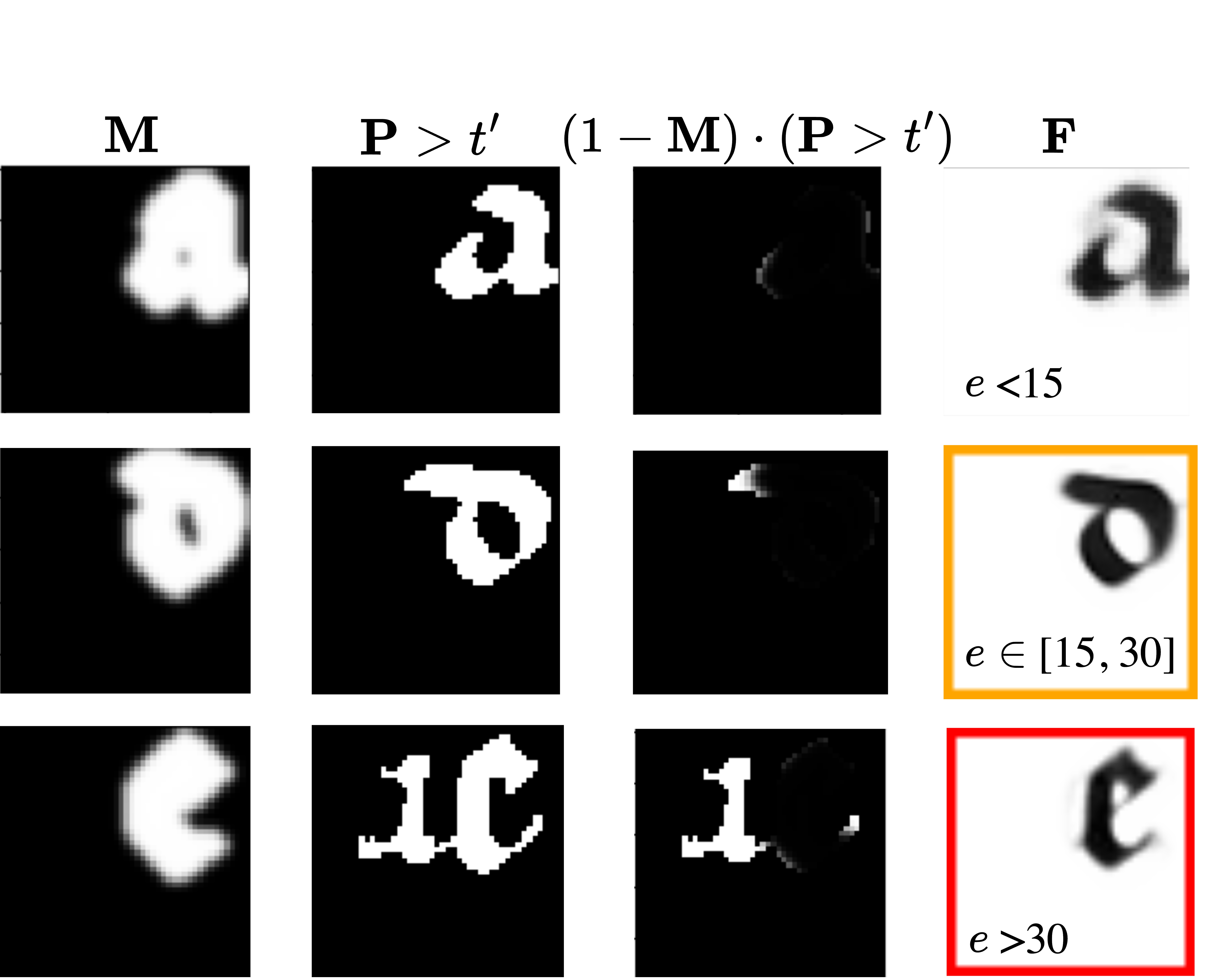}
        \caption{\textbf{Failure case identification}}
        \label{fig:graph-border-colors}
    \end{subfigure}
    \caption{\textbf{Prototype filtering and failure case identification. } We use a mask M defined from the reference prototype $\textbf{R}$ to remove artifacts from finetuned prototypes $\textbf{P}$, yieling a filtered prototype $\textbf{F}$. We compute an error $e$ associated to the filtering to automatically identify potential failure cases.}
    \label{fig:denoising-and-border-colors}
\end{figure}

While the reference prototypes generally are of high quality, we observed various artifacts in the finetuned prototypes, particularly for less common characters, when trained on a single document. While this does not hinder the qualitative analysis of the prototypes' shapes, it does complicate the quantitative comparison between prototypes. To alleviate this issue, we propose to filter the finetuned prototypes using the reference ones, as visualized in Figure~\ref{fig:denoising-and-border-colors}. \\

Let us consider a specific character, the associated reference prototype $\textbf{R}$ and the associated finetuned prototype $\textbf{P}$ for a given finetuning. Using the reference prototype, we define a reference mask $\textbf{M}$ as

\begin{equation}\label{eq:reference_masks}
\textbf{M}= \textbf{G} \ast {D} (\textbf{R} > t), 
\end{equation}
where $\textbf{G}$ is a Gaussian filter, $\ast$ denotes a convolution, $D$ is a dilation operation and $\textbf{R}> t$ is the binary mask associated to pixels for which $\textbf{R}$ is greater than a threshold $t$. In our experiments, we use a Gaussian $\textbf{G}$ of standard deviation 2, a dilation $D$ of 2 pixels, and a threshold $t=$0.8. Intuitively, this mask defines in a soft way, for each character, pixels that are close to the reference prototype.

Using this mask, we define a filtered prototype $\textbf{F}=\textbf{M}\cdot\textbf{P}$, where $\cdot$ is the pixel-wise multiplication, which we use for all of our analyses. 

\paragraph{Automatic identification of failure cases.}
While the filtering process described above generally improves the visual quality of the prototypes without changing the appearance of the characters themselves, there are instances where either the appearance is slightly altered or the finetuned prototype is of very low quality. We want to identify such cases automatically, to avoid misinterpretations. To do so, for a given character associated with a reference mask $\textbf{M}$ and a finetuned prototype $\textbf{P}$ we compute the error $e$ defined by:

\begin{equation}\label{eq:color_border_loss}
e = \| (1 - \textbf{M}) \cdot (\textbf{P} > t') \|,
\end{equation}
where $\|\ \|$ is the norm of an image, $\cdot$ is the pixel-wise multiplication, and $t'$ is a scalar threshold, set to 0.65 in our experiments. Intuitively, this error can be interpreted as the number of pixels that are present in the finetuned prototype $\textbf{P}$ (i.e., have values higher than $t'$) but are filtered out by the mask $M$. This value $e$ enables us to identify: (i) finetuned prototypes whose shape is significantly different from the reference one and are thus modified by the filtering process, such as the \character{d} in Figure~\ref{fig:denoising-and-border-colors}, and (ii) finetuned prototypes of low quality that might not be easily interpretable, such as the \character{e} in Figure~\ref{fig:denoising-and-border-colors}. In our results, we highlight prototypes where $e>15$ in orange and prototypes where $e>30$ in red.

\subsection{Character prototype comparison for palaeographic analysis}

\paragraph{Visual comparison.} We can visually highlight the morphological differences between two prototypes by subtracting one from the other. To make this difference easier to understand, we use a colormap that represents zeros as white, and positive and negative values as two distinct colors, typically red and blue. This method reveals pixel-wise differences, facilitating an initial qualitative examination of the morphological disparities (see Table ~\ref{tab:comp-litterature} and Figure~\ref{fig:visual-comp}).

\begin{figure}[t]
    \centering
    \begin{subfigure}[t]{0.31\textwidth}
    \includegraphics[width=\textwidth]{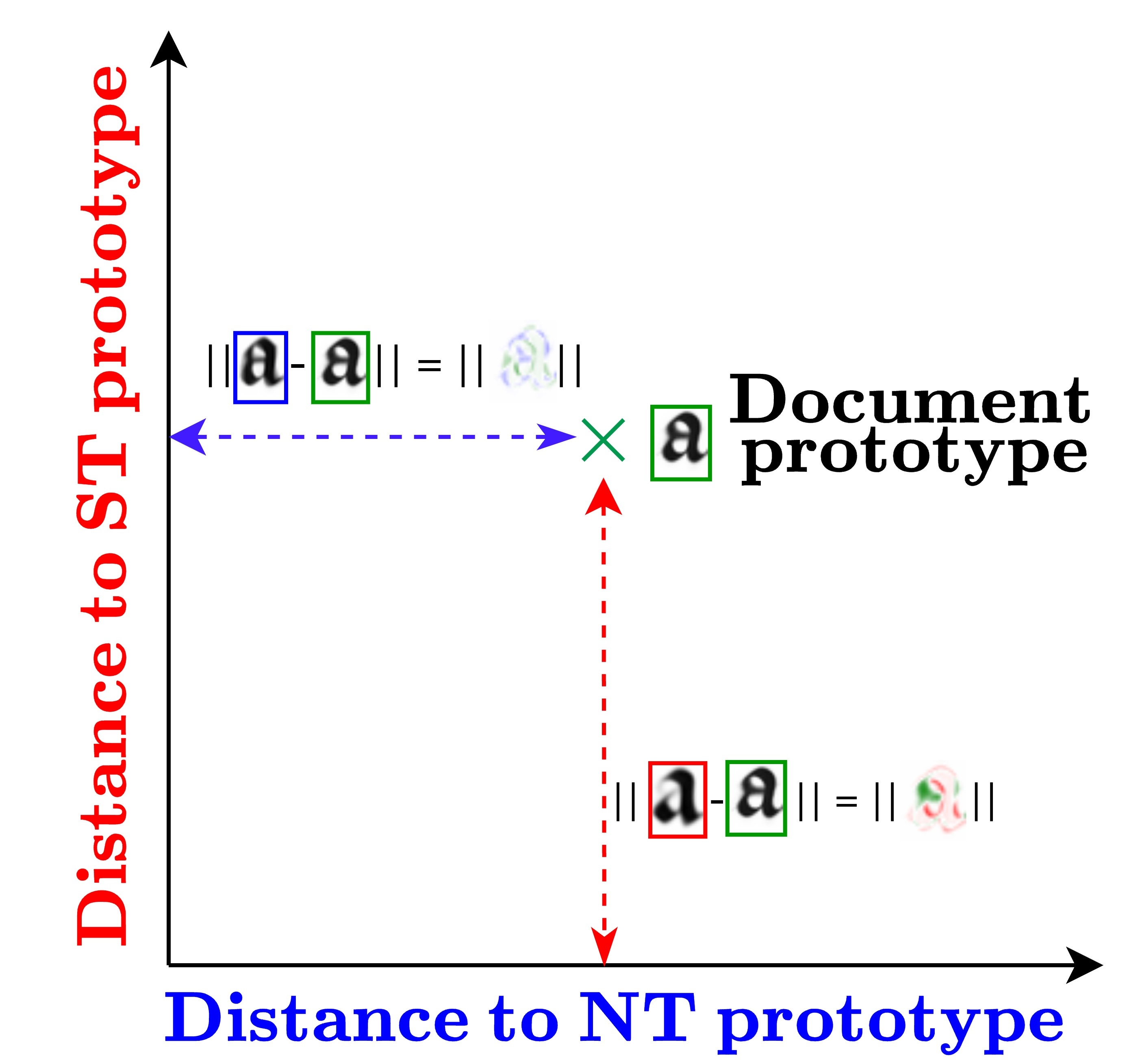}
        \caption{Graph interpretation}
        \label{fig:concept-graph-method}
    \end{subfigure}
    \begin{subfigure}[t]{0.30\textwidth}
        \includegraphics[width=\textwidth]{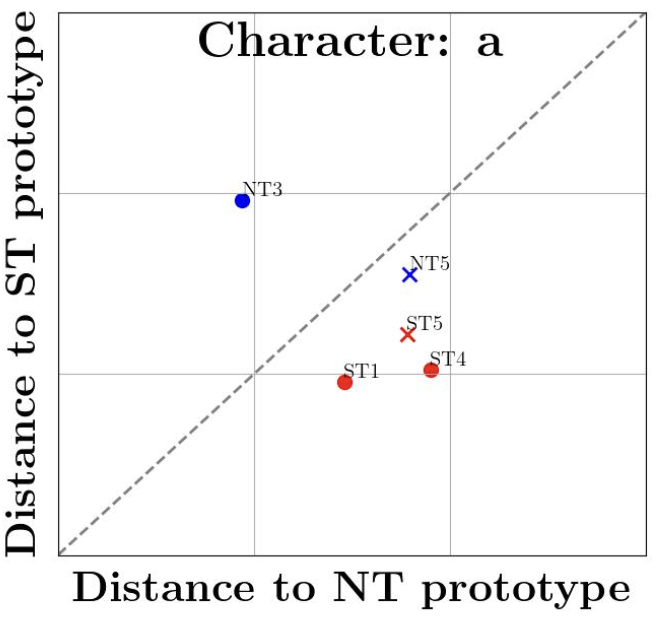}
        \caption{Character graph}
        \label{fig:example-method-a}
    \end{subfigure}
    \begin{subfigure}[t]{0.299\textwidth}\includegraphics[width=\textwidth]{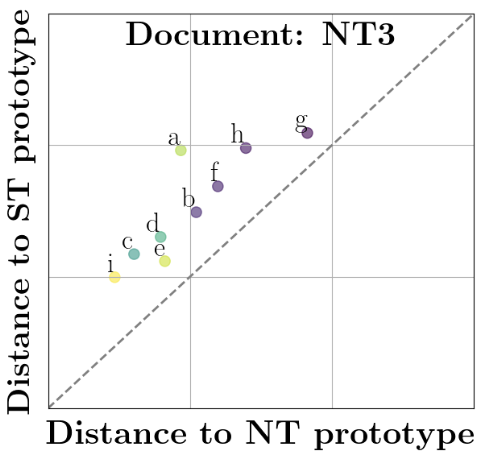}
        \caption{Document graph}
        \label{fig:example-method-manuscript}
    \end{subfigure}
        \begin{subfigure}[t]{0.05\textwidth}
    \raisebox{0.4em}{\includegraphics[width=\textwidth]{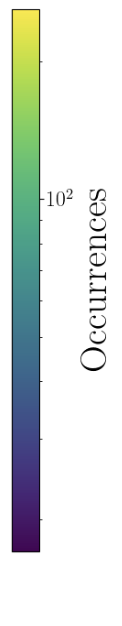}
    }
    \end{subfigure}
    \caption{\textbf{Comparison graphs.} The markers correspond to different document prototypes and their coordinates to their distance to the Northern and Southern {\it Textualis} prototypes. See text for details.}
    \label{fig:graph-method}
\end{figure}

\paragraph{Character and document comparison graphs.} To quantitatively analyze character prototypes, we introduce an adapted comparison graph, illustrated in Figure~\ref{fig:graph-method}. In this graph, each point represents a specific document character prototype, with its coordinates defined as its distance in pixel space to two selected prototypes. 
Since we study \textit{Textualis Formata}, we use the prototypes of Northern and Southern \textit{Textualis} (NT and ST) (see Section~\ref{sec:data} for details), other prototypes could be selected for different analysis. The distance to the axes can be interpreted as visualized in Figure~\ref{fig:concept-graph-method}.

We employ two complementary types of graphs for our analysis. Firstly, \textit{character graphs} (see Figure~\ref{fig:example-method-a}) concentrate on a single character across all documents. In these graphs, dots represent the documents selected to train the reference, Northern and Southern \textit{Textualis} models,
while crosses denote the remaining documents. 
Blue, resp. red, markers signify Northern, resp. Southern \textit{Textualis} documents. The identifier for each document is written near its marker. 
This type of graph allows to easily identify outlier documents for a specific character, such as NT5 for the \character{a} character, which is closer to the ST prototype. 
Secondly, \textit{document graphs} (see Figure~\ref{fig:example-method-manuscript}) focus on a specific document, where each dot corresponds to a different character, labeled near the marker. The color of the dots corresponds to the frequency of occurrence, with darker dots representing less frequent characters. This type of graph facilitates the identification, for a given document, of the characters most typical of a subtype, such as the \character{a} for NT3.

\paragraph{Quantifying character variability.}\label{sec:variability} According to the literature, the Northern \textit{Textualis} class allows for more morphological variation across documents than the Southern \textit{Textualis}. While the visualizations and graphs can qualitatively support this idea, we further aim to quantify the characters' variability. Thus, we report the standard deviations of character prototypes within one subtype, $\sigma_{NT}$ for Northern \textit{Textualis} and $\sigma_{ST}$ for Southern \textit{Textualis}. These standard deviations can be thought of as the average number of pixels that change across two character prototypes of the same subtype.

\section{Experiments}

\paragraph{Research question and analysis framework.} To analyze the results of our approach, we adopt the taxonomy formalized by A. Derolez~\cite{derolez2003palaeography}, which provides a framework based on morphological criteria. We select a corpus in \textit{Textualis} script type, specifically in its canonized calligraphic form \textit{Formata}, due to its more distinguishable morphological elements compared to more rapidly executed forms. Despite the morphology-based categorization where \textit{Textualis Formata} represents a coherent group, Derolez
makes \enquote{an important distinction between two fundamentally different species}, Northern and Southern \textit{Textualis}, based on their geographical distribution and a set of minute morphological differences. However, these differences vary according to factors such as date, geographical origin, or language, and often intersect, blurring this fundamental distinction in some cases. Our goal is to confront the results of our approach with Derolez’s criteria and observations.

\begin{table}[t]
    \centering
    \scalebox{0.96}{}
    \caption{\textbf{Dataset Description. }The `Doc.' column refers to the names we use for the different documents in this paper, NT and ST stand for Northern and Southern \textit{Textualis}, and the `Ref.' column reports which documents are used to train the reference and subtype models.}
    \label{tab:dataset_description}
    \resizebox{\linewidth}{!}{
    \begin{tabular}{c c c c c c c c c c}
    \textbf{Doc.} & \textbf{Ref.} & \textbf{Shelfmark} & \textbf{Language} & \textbf{Century}& \textbf{Date}&  \textbf{Origin} &\textbf{Folio(s)} &\textbf{Lines}  \\ \hline 
    NT1 & \checkmark & Paris, BnF, Français 403~\cite{ecmen2017stutzmann}   & French  & 13\textsuperscript{th} & 1226-1250 & England & 4r   & 76  \\
    NT2 & \checkmark & Paris, BnF, Français 12400~\cite{ecmen2017stutzmann} & French    & 14\textsuperscript{th} & 1305-1310 & Eastern France& 92r  & 54  \\
    NT3 & \checkmark &  Arras, BM Ms. 861 (315)~\cite{CREMMA_Medii_Aevi}  & Latin    & 14\textsuperscript{th} &     -      & - & 56r  & 65 \\
    NT4 & \checkmark &Paris, BnF, Français 1728~\cite{Cremma_Medieval}      & French   & 14\textsuperscript{th} &  1372     & - & 3r  & 59 \\
    NT5 &  &Paris, BnF, Français 20120~\cite{ecmen2017stutzmann}  & French    & 13\textsuperscript{th} & 1240-1250 & Paris or Orleans&  7r & 81  \\
    NT6 &  &Paris, BnF, Français 619~\cite{HTRomance_Med_French}  & French    & 14\textsuperscript{th} &1375-1400  & - & 1v   & 65  \\
    NT7 &  &Berlin, SB, Hdschr. 25~\cite{CREMMA_Medii_Aevi}       & Latin    & 15\textsuperscript{th} &    1451-1500       & Flanders & 22r-22v & 25  \\ \hline
    ST1 & \checkmark &Paris, BnF, Français 9082~\cite{ecmen2017stutzmann}   & French   & 13\textsuperscript{th} & 1295      &Rome& 171r & 56   \\
    ST2 & \checkmark &Paris, BnF, Espagnol 65~\cite{HTR_Medieval_Spain}     &Navarrese  & 14\textsuperscript{th} & 1301-1310    & - & 5v,6r& 118  \\
    ST3 & \checkmark &Paris, BnF, Italien 590~\cite{HTR_Medieval_Italian}   & Italian   & 14\textsuperscript{th} & 1370-1410 &Italy& 18r  & 68   \\
    ST4 & \checkmark &Madrid, FLG, mss 289 (Hand A)~\cite{HTR_Castilan}              &Castilian  & 15\textsuperscript{th} & 1480      & Seville& 245v & 86   \\
    ST5 &  &Paris, BnF, Français 187~\cite{ecmen2017stutzmann}    & French  & 14\textsuperscript{th} & 1350-1386 & Milan or Genova & 18r  & 16   \\
    ST6 &  &Paris, BnF, Latin 7720~\cite{HTR_Medieval_Latin}      & Latin     & 15\textsuperscript{th} & 1390-1410 & Florence& 102v & 74   \\
    ST7 &  &Madrid, FLG, mss 289 (Hand B)~\cite{HTR_Castilan}              &Castilian  & 15\textsuperscript{th} & 1480      & Seville& 274v & 52  \\ \hline
    Total:         &      &   &                     &                        &  &   &   &  \textbf{892}   \\        
        \end{tabular}}
\end{table}

\subsection{Dataset and Experiment Details}
\label{sec:data}
\paragraph{Data Selection.} Our data was build from two open-access repositories, ECMEN~\cite{stutzmann__2017,ecmen2017stutzmann} and CATMuS~\cite{pinche2023catmus}. We selected the documents to ensure the variability of the corpus in terms of geographic, linguistic, and chronological distribution, resulting in seven documents for each subscript as listed in Table~\ref{tab:dataset_description} (labelled NT 1-7 and ST 1-7). We verified and normalized the transcriptions to fit a graphemic approach~\cite{Clerice_Choco-Mufin_a_tool_2021,CREMMA_Medii_Aevi}. Despite our efforts to use diverse and representative documents, we acknowledge that biases may exist within our dataset.

\begin{figure}[t!]
  \centering
    \includegraphics[width=\textwidth]{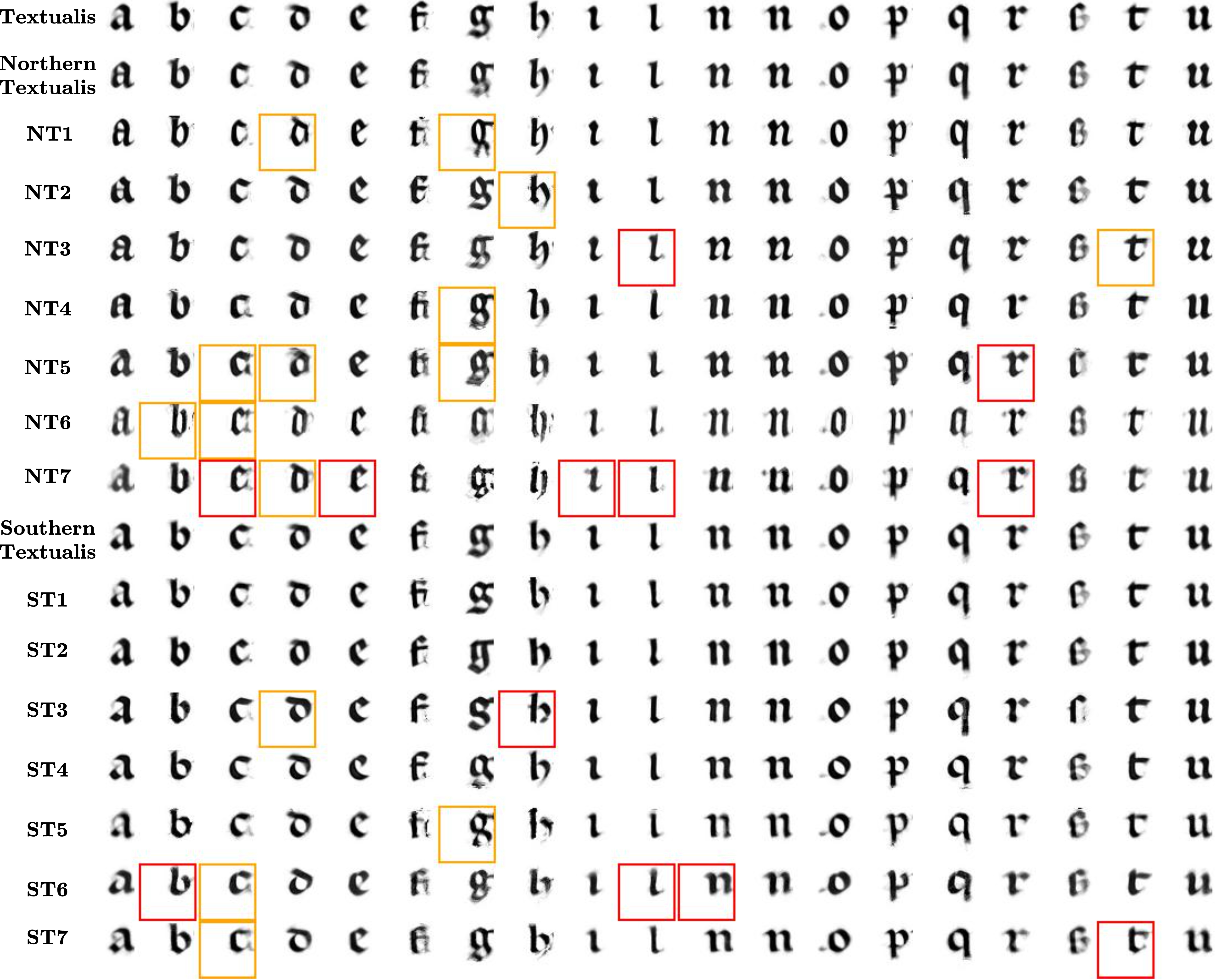}
    \caption{\textbf{Our filtered prototypes} on type, sub-type and document level. The highlighted prototypes are the ones for which filtering had a significant impact (see Section~\ref{sec:ltw} and~\ref{sec:res} for details).}
        \label{fig:denoised-borders}
\end{figure}

\paragraph{Character set choice.} 
\label{sec:res}
From the extended set of characters in medieval manuscripts - including upper and lower case letters, ligatures, punctuation, and abbreviation signs - we follow a standard approach for morphological analysis and focus on the lowercase alphabetic characters where morphology is crystalized through frequent usage. From these characters, we show results on the ones common to all documents, thus excluding \character{j,k,v,x,y,z}.

\paragraph{Trained models.} For our analysis, we trained multiple models to obtain character prototypes at different levels of granularity: (i) a script type model for \textit{Textualis}, (ii) script subtype models for Northern and Southern \textit{Textualis}, and (iii) document level models for each document in our dataset. We use the \textit{Textualis} script type model as reference model, and finetune all other models from it as explained in Section~\ref{sec:ltw}. to validate that our reference and subtype models can be effective to analyze documents they were not trained on, we limited their training to NT 1-4 and ST 1-4, indicated in the `Ref' column in Table~\ref{tab:dataset_description}.

\subsection{General results}

\paragraph{Prototype quality.} Figure~\ref{fig:denoised-borders} shows the character prototypes generated by our approach across type, sub-type, and document levels. We note several limits in these prototypes. Firstly, the prototype for \character{m} is not well modeled (\inlinegraphics{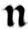}). This issue stems from the network being trained with a CTC loss allowing to use the same prototype twice to model the same letter. As a result, we exclude \character{m} from our analysis. Secondly, the two allographs \character{\longs/ s}
are represented by a single prototype 
resulting in an averaged representation of the two (\inlinegraphics{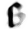}), necessitating cautious examination. Thirdly, we meticulously scrutinized all prototypes highlighted in orange and red, where filtering significantly impacted the outcomes. In almost all instances, the filtering was meaningful and eliminated irrelevant artifacts from the prototypes. Nevertheless, it is worth mentioning that:
(i) the lengthy shaft of the ST3 \character{d}(\inlinegraphics{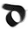}) and the hairline extension of the limb of \character{h} in NT2 (\inlinegraphics{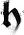}) are slightly severed; (ii) in general, documents that are not part of the reference set, are less accurately modeled and more impacted by the filtering, especially the forked ascender tops of \character{h} and \character{b}. This information loss, while limited, should be considered during the analysis.

\afterpage{%
\newcolumntype{C}[1]{>{\centering\arraybackslash}p{##1}}

\newcolumntype{L}[1]{>{\raggedright\arraybackslash}p{##1}}

\begin{longtable}[ht!]{ C{0.9cm}|p{9cm}|L{2.1cm}} 

\caption{Comparison between Derolez' criteria for Northern and Southern \textit{Textualis} and our subtype prototypes.}
\label{tab:comp-litterature} \\
\textbf{{\character{Ch.}}} &
\centering\textbf{Derolez'criteria} &  \textbf{NT\hspace{0.02cm}|\hspace{0.03cm}ST\hspace{0.03cm}|diff.}   \\[-0.5em]
 & & \vspace{-1em}\scriptsize{$\sigma_{NT}$} \ \scriptsize{$\sigma_{ST}$} \\
\hline

\textbf{    \character{a}}  & \textbf{NT}: Closed form with variations like \enquote{box-\character{a}} \newline \textbf{ST}: Open form or slightly closed with hairline  &
  \raisebox{-1.28em}
{\centering\includegraphics[height=2em]{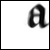}\hspace{-1mm} \includegraphics[height=2em]{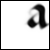}\hspace{-1mm} \includegraphics[height=2em]{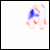}}\vspace{-0.7em}
\scriptsize{4.0 \hspace{2mm} 3.4}
\\
\hline
\textbf{    \character{b}}   & \textbf{NT}: Sloped or forked ascender tops \newline \textbf{ST}: (i) Flat ascender tops, (ii) round lobe &  
\raisebox{-1.28em}
{\centering\includegraphics[height=2em]{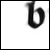}\hspace{-1mm} \includegraphics[height=2em]{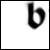}\hspace{-1mm}
\includegraphics[height=2em]{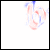}}\vspace{-0.7em}
\scriptsize{ 4.1 \hspace{2mm} 3.6} \\
\hline

\textbf{    \character{c}}   &  \textbf{NT}: Angular or broken lobe curves  \newline \textbf{ST}: Semi-circular lobe  
&
\raisebox{-1.28em}
{\centering\includegraphics[height=2em]{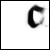}\hspace{-1mm} \includegraphics[height=2em]{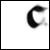}\hspace{-1mm}
\includegraphics[height=2em]{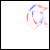}}\vspace{-0.7em}
\scriptsize{ 2.9 \hspace{2mm} 2.4 } \\
\hline

\textbf{    \character{d}} & \textbf{NT}: (i) Lengthened  and (ii) concave shaft \textbf{ST}: (i) Shorter shaft and (ii) almost horizontal, (iii) round bowl & 
  \raisebox{-1.28em}
{\centering\includegraphics[height=2em]{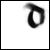}\hspace{-1mm} \includegraphics[height=2em]{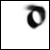}\hspace{-1mm}
\includegraphics[height=2em]{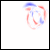}}\vspace{-0.7em}
\scriptsize{ 3.8 \hspace{2mm} 3.1 } \\
\hline
\textbf{    \character{e}} & \textbf{NT}: (i) Diagonal direction of the hairline and (ii) angular or broken lobe curves \newline \textbf{ST}: (i) Horizontal or no hairline,  (ii) semi-circular lobe form & 
  \raisebox{-1.28em}
{\centering\includegraphics[height=2em]{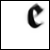}\hspace{-1mm} \includegraphics[height=2em]{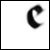}\hspace{-1mm}
\includegraphics[height=2em]{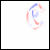}}\vspace{-0.7em}
\scriptsize{ 3.3 \hspace{2mm} 3.2 } \\
\hline
\textbf{    \character{f}}  & \textbf{NT}: Incurvation of the shaft foot to the right  \newline \textbf{ST}: Flat foot
& \raisebox{-1.28em}
{\centering\includegraphics[height=2em]{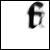}\hspace{-1mm} \includegraphics[height=2em]{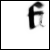}\hspace{-1mm}
\includegraphics[height=2em]{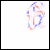}}\vspace{-0.7em}
\scriptsize{ 4.3 \hspace{2mm} 4.3} \\
\hline

\textbf{    \character{g}}  &  \textbf{NT:} Tendency for the closed, \enquote{8-shaped} form \newline \textbf{ST}: Tendency for the open, \enquote{Rücken -g} form \newline\textbf{Note}:  Various intermediate forms and difficult to classify &
  \raisebox{-1.28em}
{\centering\includegraphics[height=2em]{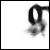}\hspace{-1mm} \includegraphics[height=2em]{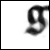}\hspace{-1mm}
\includegraphics[height=2em]{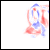}}\vspace{-0.7em}
\scriptsize{ 5.8 \hspace{2mm} 5.4 }  \\
\hline

\textbf{    \character{h}}  & \textbf{NT}: (i) Incurvation of the shaft foot to the right, (ii) extended or dislocated limb and (iii) sloped or forked tops \textbf{ST}: (i) Flat ascender foot, (ii) circular limb on the baseline and (iii) flat ascender tops & 
\raisebox{-1.28em}
{\centering\includegraphics[height=2em]{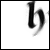}\hspace{-1mm} \includegraphics[height=2em]{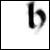}\hspace{-1mm}
\includegraphics[height=2em]{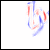}}\vspace{-0.7em}
\scriptsize{ 4.9\hspace{3mm} 4.2 } \\
\hline
\textbf{    \character{i}}  & \textbf{NT}: (i) Accentuated (diamond-shaped or forked) headline and (ii) extended hairline for the foot \textbf{ST}: (i) Approach stroke for the headline and (ii) flat end for the foot & 
\raisebox{-1.28em}
{\centering\includegraphics[height=2em]{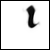}\hspace{-1mm} \includegraphics[height=2em]{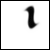}\hspace{-1mm}
\includegraphics[height=2em]{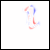}}\vspace{-0.7em}
\scriptsize{ 2.4 \hspace{3mm}1.6 } \\
 \hline
\textbf{    \character{l}}  & \textbf{NT}: Sloped or forked ascender tops \newline \textbf{ST}: Flat tops & 
  \raisebox{-1.28em}
{\centering\includegraphics[height=2em]{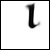}\hspace{-1mm} \includegraphics[height=2em]{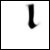}\includegraphics[height=2em]{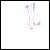}}\vspace{-0.7em}
\scriptsize{ 2.5 \hspace{2mm} 1.8 }\\
\hline
\textbf{    \character{n}}  & \textbf{NT}: Accentuated (diamond-shaped or hairlines) for the headline and (ii) same for feet \textbf{ST}: (i) Approach stroke hairline for the headline and (ii) flat feet  &  
\raisebox{-1.28em}
{\centering\includegraphics[height=2em]{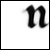}\hspace{-1mm} \includegraphics[height=2em]{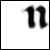}\hspace{-1mm} \includegraphics[height=2em]{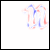}}\vspace{-0.7em}
\scriptsize{ 3.7\hspace{3mm} 3.0} \\
\hline

\textbf{    \character{o}}  & \textbf{NT}: Broken / more vertically elongated curves \newline \textbf{ST}: Circular arc forms  &   
\raisebox{-1.28em}
{\centering\includegraphics[height=2em]{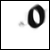}\hspace{-1mm} \includegraphics[height=2em]{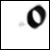}\hspace{-1mm} \includegraphics[height=2em]{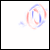}}\vspace{-0.7em}
\scriptsize{ 3.5 \hspace{2mm} 2.7}  \\
\hline

\textbf{    \character{p}}  & \textbf{NT}: (i) Artificial spurs on the left and (i) decorated descender feet \textbf{ST}: No spurs and (ii) flat descender feet &

\raisebox{-1.28em}
{\centering\includegraphics[height=2em]{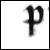}\hspace{-1mm} \includegraphics[height=2em]{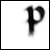}\hspace{-1mm} \includegraphics[height=2em]{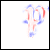}}\vspace{-0.7em}
\scriptsize{ 4.2 \hspace{2mm} 3.2 }  \\
\hline

\textbf{    \character{q}}  & \textbf{NT}: (i) Lengthy and (ii) decorated descenders \newline \textbf{ST}: (i) Short and (ii) flat descenders &

  \raisebox{-1.28em}
{\centering\includegraphics[height=2em]{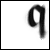}\hspace{-1mm} \includegraphics[height=2em]{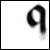}\hspace{-1mm} \includegraphics[height=2em]{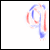}}\vspace{-0.7em}
\scriptsize{ 4.8 \hspace{2mm} 4.0} \\
\hline
\textbf{    \character{r}}   & \textbf{NT}: (i) Hairline endstroke for shaft foot and (ii) angular horizontal stroke \newline \textbf{ST}: (i) Flat shaft foot and (ii) straight horizontal stroke &
 \raisebox{-1.28em}
{\centering\includegraphics[height=2em]{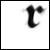}\hspace{-1mm} \includegraphics[height=2em]{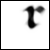}\hspace{-1mm} \includegraphics[height=2em]{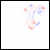}}\vspace{-0.7em}
\scriptsize{ 2.8 \hspace{2mm} 2.4 } \\
\hline
\textbf{    \character{s}}   & \textbf{NT}: Incurvation of the shaft foot to the right  for \character{\longs} and (ii) closed and angular curves for \character{s} \textbf{ST}: (i) Flat shaft foot for \character{\longs} and (ii) open semi-circular curves for \character{s} &  
\raisebox{-1.28em}
{\centering\includegraphics[height=2em]{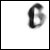}\hspace{-1mm} \includegraphics[height=2em]{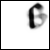}\hspace{-1mm} \includegraphics[height=2em]{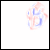}}\vspace{-0.7em}
\scriptsize{ 2.7\hspace{3mm} 2.4}\\
\hline
\textbf{    \character{t}}  & \textbf{NT}: Vertical pendant hairline of the headstroke \textbf{ST}: No ornaments  \textbf{Note}: Different levels of shaft projection above headline and length of horizontal stroke   &
  \raisebox{-1.28em}
{\centering\includegraphics[height=2em]{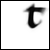}\hspace{-1mm} \includegraphics[height=2em]{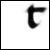}\hspace{-1mm} \includegraphics[height=2em]{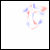}}\vspace{-0.7em}
\scriptsize{ 3.0 \hspace{2mm} 2.4} \\ 
\hline

\textbf{    \character{u}} & \textbf{NT}: Accentuated (diamond-shaped or sloped) headline \newline \textbf{ST}: Flat or left approach stroke for headline &
  \raisebox{-1.28em}
{\centering\includegraphics[height=2em]{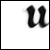}\hspace{-1mm} \includegraphics[height=2em]{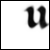}\hspace{-1mm} \includegraphics[height=2em]{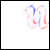}}\vspace{-0.7em}
\scriptsize{ 3.7\hspace{3mm} 3.0} \\
\hline
\end{longtable}
}

\paragraph{Palaeographical relevance of the subtype prototypes.} To showcase how our prototypes can be related to classical palaeographic analysis, we systematically compare in Table~\ref{tab:comp-litterature} Derolez's general morphological criteria to our Northern and Southern \textit{Textualis} prototypes, highlighting their variations by visualizing their difference. We find that Derolez's observations closely align with the variations that our prototypes enable us to visualize. Additionally, we report our variability scores $\sigma_{NT}$ and $\sigma_{ST}$ for each letter, which were consistently higher for Northern \textit{Textualis}, which is consistent with Derolez's claim that this script subtype generally exhibits higher intra-class variation.

\subsection{Character graph analysis.}
\label{letter-analysis}
\begin{figure}[t!]      \includegraphics[width=\linewidth]{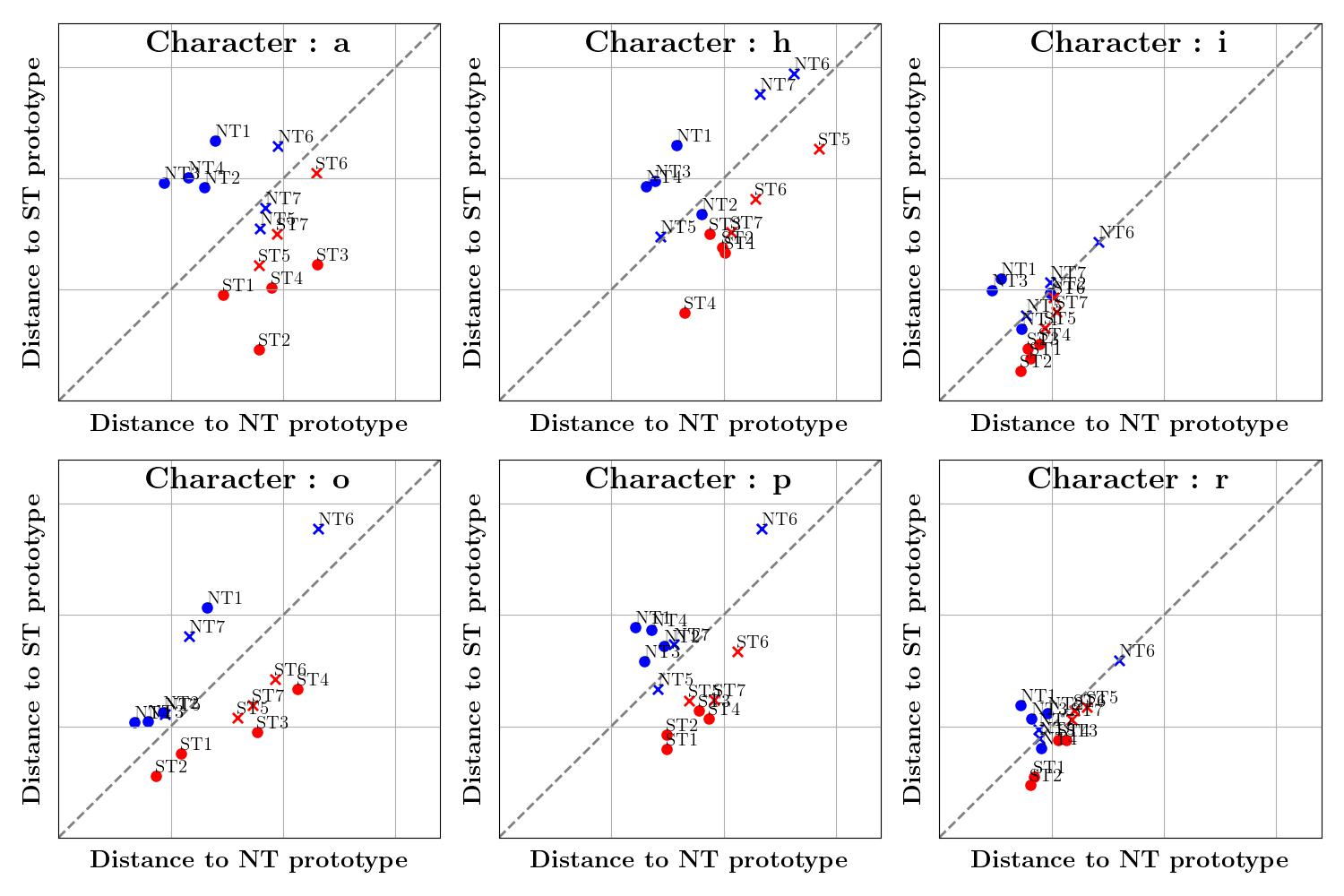}
    \caption{\textbf{Character comparison graphs}.
    }
        \label{fig:letter-graphs}
    \end{figure}

In this section, we provide examples of how our character graphs, together with our prototypes, can support a detailed palaeographic analysis of the variations of a specific character (examples presented in Figure~\ref{fig:letter-graphs}).

\paragraph{Discriminative characters.} We first analyze the results for four discriminative characters, \character{a,o,p,h}. The letter \textbf{\character{a}} is often considered as a distinguishing criterion between script types, so much so that W. Oeser~\cite{oeser1971grundlage} distinguished seven categories within the Northern \textit{Textualis} script subtype mainly based on allographs of \character{a}. Most striking in our \textbf{\character{a}} character graph is that the prototypes for NT5 (\inlinegraphics{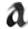}) and NT7 (\inlinegraphics{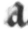}) are actually closer to the ST prototypes. This is consistent with the observation that open \character{a} forms are standard for ST. The dispersion of the characters on the graph also provides insight into the variability of \character{a} in this subtype. The group associated to NT1-4 corresponds to the closed 
\enquote{box-a} form in NT2 (\inlinegraphics{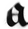}) and NT4 (\inlinegraphics{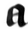}) and the double-bow variant in NT1 (\inlinegraphics{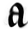}) and NT3 (\inlinegraphics{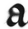}). NT6 presents a more vertically elongated form and stands out (\inlinegraphics{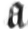}). While there is morphological variations across ST documents, with round shapes (ST1 \inlinegraphics{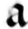}; 
ST2 \inlinegraphics{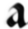};
ST5 \inlinegraphics{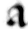}; 
ST6 \inlinegraphics{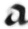}), or with more angular inner bows (ST3 \inlinegraphics{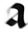};
ST4 \inlinegraphics{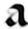};
ST7 \inlinegraphics{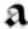}), the consistent use of an open form, or only closed with a hairline, distinguishes them from the NT subtype, and all ST documents prototypes are closer to the ST prototype.

The letter \textbf{\character{o}} is also particularly discriminative. The treatment of its (generally) two mirroring arcs, using broken or semi-circular strokes, often has visual echoes in letters with lobes and arcs like \character{b, c, e, p, q}, which contributes to the visual evaluation of a hand or script type as wide/round or narrow/angular. This is confirmed by the fact that all the points of documents identified as NT are above the diagonal and all the ones corresponding to ST below, meaning that the prototypes for each document are closer to its subtype prototype than the other one. For NT, the forms of NT2-5 (\inlinegraphics{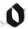}\inlinegraphics{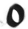}\inlinegraphics{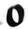}\inlinegraphics{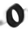}) are particularly well reconstructed, better than NT1 (\inlinegraphics{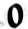}) and NT7 (\inlinegraphics{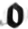}) which are slightly more narrow and vertically elongated. Again, the form of NT6 stands out (\inlinegraphics{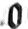}), it consists of double broken strokes resulting in a narrow, quadrangle shape. Similarly for ST, ST1 (\inlinegraphics{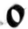}) and ST2 (\inlinegraphics{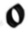}) are particularly close to the ST prototype, being less wide than ST3-7 (\inlinegraphics{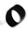}\inlinegraphics{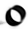}\inlinegraphics{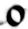}\inlinegraphics{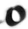}\inlinegraphics{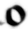}). This is consistent with Derolez' assertion that angularity/roundness separates the two subtypes, while intra-class variation is associated with different degrees of narrowness/breadth.

Regarding the letter \textbf{\character{h}}, it highlights one of the limitations of our approach. The extended limb, characteristic of NT (cf. Table~\ref{tab:comp-litterature}) is clearly present in all associated documents and prototypes. However, because its position varies in a document, the limb appears dimmed in the NT prototype (\inlinegraphics{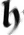}). Moreover, its position varies significantly across different documents, which can result in a greater distance between a document prototype and the NT prototype. This explains why the NT2 prototype is actually more similar to the ST prototype, while NT5 is as close to both. Note that the shift of the limb in NT2 (not curved to the left but rather extended straight down) compared to the NT prototype is so significant that our filtering partially erases it (as can be confirmed by examining the masked region, similar to Figure~\ref{fig:graph-border-colors}), which was flagged by our automatic failure identification. This emphasizes the necessity to confront our graphs with the visual appearance of the prototypes and the documents for interpretation.

For the letter \textbf{\character{p}}, we note that the consistent presence of artificial spurs at the baseline level and in general of decorations are characteristic of NT (\inlinegraphics{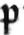}). The two subtypes appear well separated, except for the deviation of NT5 (\inlinegraphics{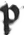}), which we will further analyze in Section ~\ref{manuscript-analysis}.

\paragraph{Non-discriminative characters.} We now examine \textbf{\character{i}} and \textbf{\character{r}}.
In our graphs, the points corresponding to these letters in all documents are close to the diagonal, i.e., they are as close to both the NT and ST prototypes. The fact that they are almost all close to the origin indicates that they present little variation. For \character{i}, NT1 and NT3 stand out as more typical of NT, and they indeed show clear diamond-shaped headlines (\inlinegraphics{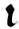}\inlinegraphics{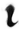}). The NT4 prototype (\inlinegraphics{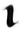}), on the other hand, is actually closer to the ST prototypes, and it does not present any headline decoration typical of NT. For both characters, the NT6 prototypes are much further than the rest from both the NT and ST prototypes and correspond to much narrower and elongated forms (\inlinegraphics{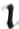} \inlinegraphics{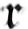}).

\subsection{Document graph analysis.}\label{manuscript-analysis}

\begin{figure}[t!]      \includegraphics[width=\linewidth]{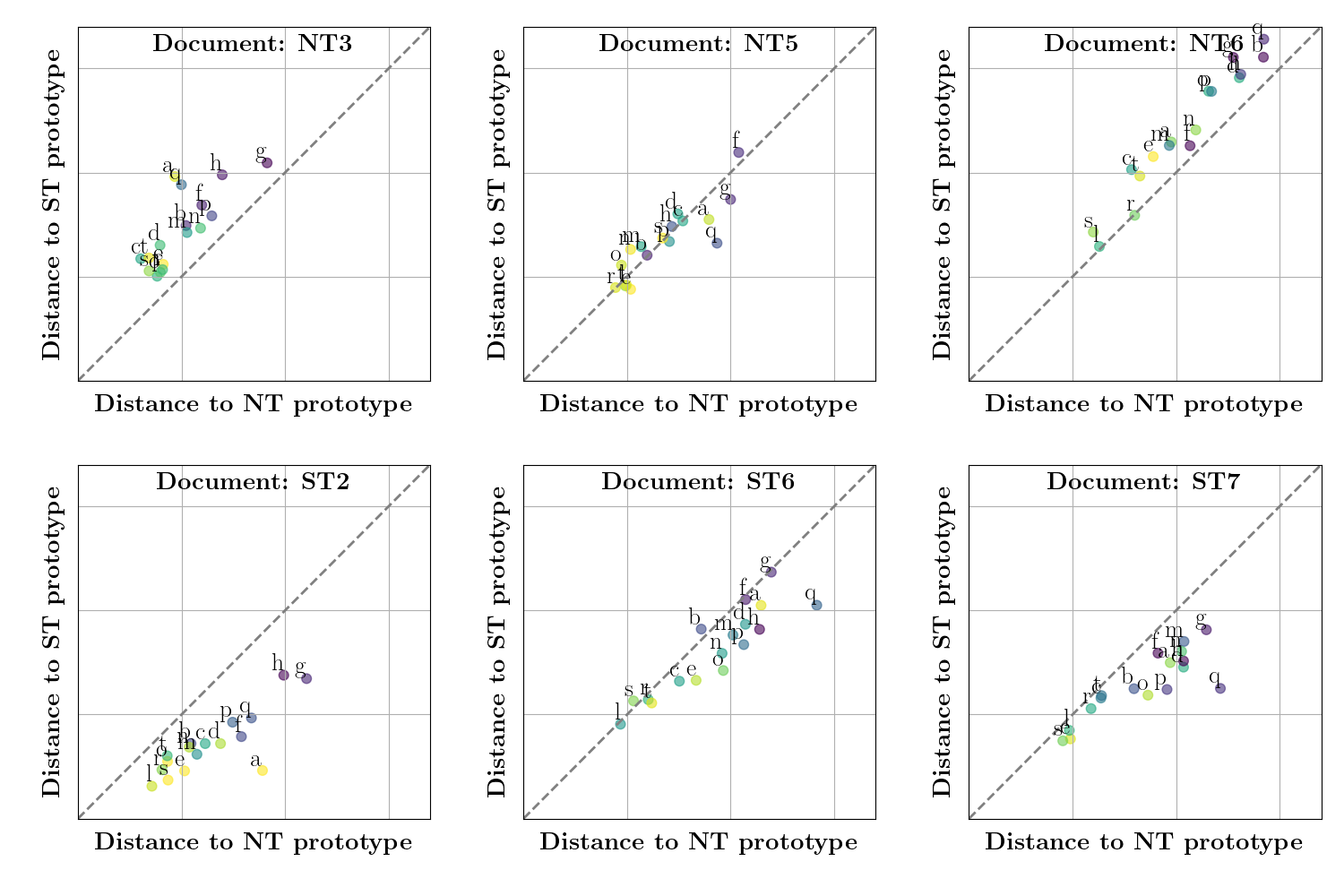}
\caption{\textbf{Document comparison graphs}.}
    \label{fig:mss-graphs}
\end{figure}

\begin{figure}[t!]
  \centering   \includegraphics[width=\textwidth]{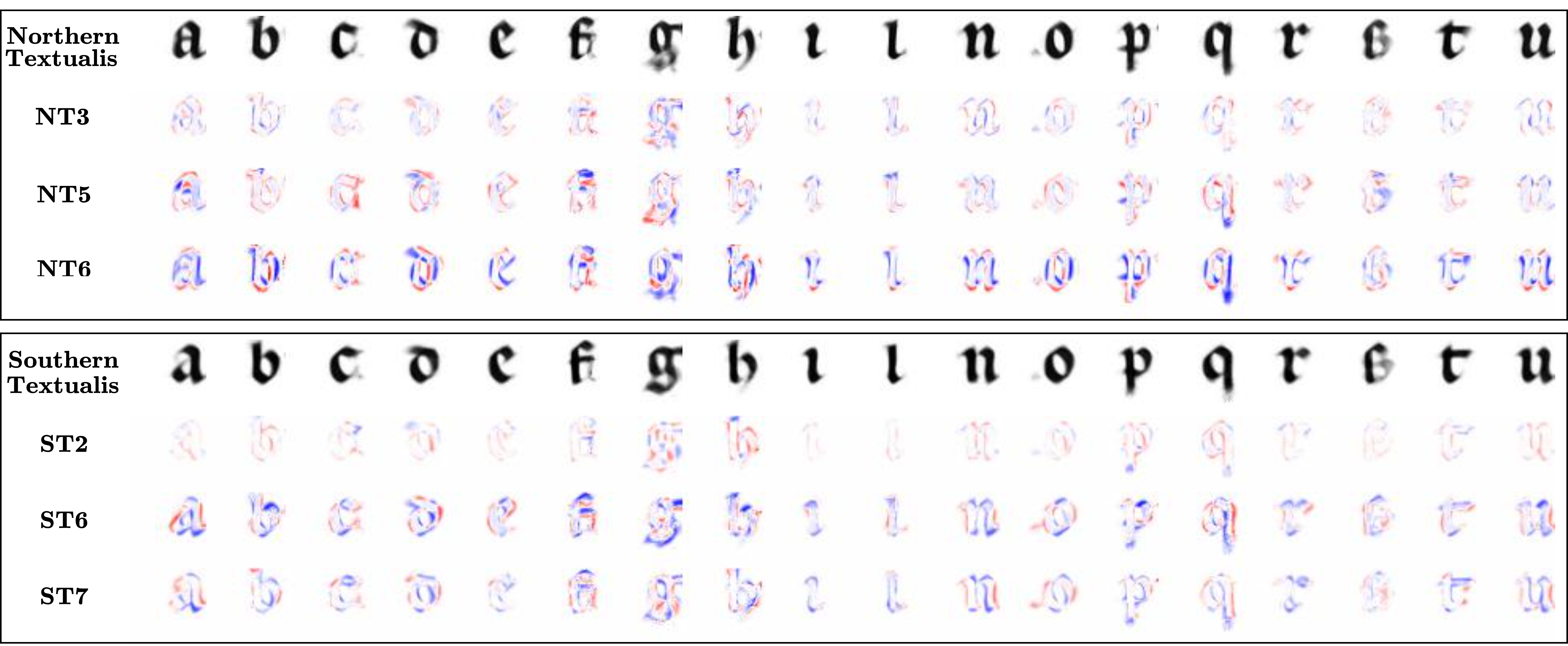}  \caption{\textbf{Visual comparison.} Subtype and document prototypes and their pixel-wise differences with positive values in blue and negative in red.}
    \label{fig:visual-comp}
\end{figure}

In this section, we discuss and interpret document graphs for the examples presented in Figure~\ref{fig:mss-graphs}. Additionally, we visualize the differences between the document prototypes and subtype prototypes in Figure~\ref{fig:visual-comp}, leveraging them to better understand the graphs.

\paragraph{Class-representative cases.} We start by examining two documents that are very typical of their subtype. For \textbf{NT3}, a 14\textsuperscript{th} century manuscript in Latin, the graph clearly shows that all the character prototypes are closer to the NT prototypes than the ST prototypes. More in details, NT3 presents a closed, double bow \character{a} (\inlinegraphics{figures/south_north/sprites_text/NT3/28.png}), which progressively dominated over other variants from the end of the 13\textsuperscript{th} c.~\cite{derolez2003palaeography}, ascenders are consistently sloped on the left side (\inlinegraphics{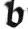}\inlinegraphics{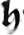}\inlinegraphics{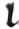}, hairlines directed to the right at the baselines (\inlinegraphics{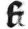}\inlinegraphics{figures/south_north/sprites_text/NT3/36.png}\inlinegraphics{figures/south_north/sprites_text/NT3/39.png}\inlinegraphics{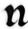}\inlinegraphics{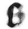}\inlinegraphics{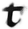}), and there are clear diamond-shaped headlines (\inlinegraphics{figures/south_north/sprites_text/NT3/36.png}\inlinegraphics{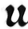}). All characters thus fully correspond to the expected NT forms. For \textbf{ST2}, copied in the first decade of 14\textsuperscript{th} c. in the Iberian peninsula, the character prototypes are, on the contrary, all closer and conforming to the ST prototypes: compact letters with very short flat-top ascenders (\inlinegraphics{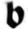}\inlinegraphics{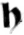}), flat-feet descenders (\inlinegraphics{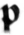}\inlinegraphics{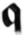}) and strokes so bold, hairlines almost become invisible (\inlinegraphics{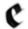}\inlinegraphics{figures/south_north/sprites_text/ST2/35.png}).

\paragraph{Ambiguous cases.} A particular interest of our document graph method is the identification and analysis of documents that partially diverge from their assigned subtype. 
\textbf{NT5} stands out in the graphs, as seven character prototypes are closer to ST than to   NT prototypes, with a particular difference for \character{a,g,q} (\inlinegraphics{figures/south_north/sprites_text/NT5/28.png}\inlinegraphics{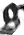}\inlinegraphics{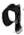}). Copied in the 13\textsuperscript{th} c., in the Paris/Orleans area, NT5's forms are significantly smaller in size, an example of \enquote{pearl-script}, generally used for Parisian pocket-Bibles. Even though its level of execution is still \textit{Formata}, due to their size, certain letters are simplified, resulting in forms that are closer to ST, like open \character{a} (\inlinegraphics{figures/south_north/sprites_text/NT5/28.png}) - a characteristic of early NT samples -, and spurless \character{p} (\inlinegraphics{figures/south_north/sprites_text/NT5/43.png}). On the contrary, a closer examination of \character{g} reveals it is in reality characteristic of NT, but not well modeled by our prototypes, in part because of the high level of variation, resulting in a blurred prototype, and in part because it was too different from the reference documents, which was actually flagged by our automatic failure case identification. \textbf{ST6}, copied in Florence at the end of 14\textsuperscript{th}/beginning of 15\textsuperscript{th} c., also presents two class diverging characters, \character{b} and \character{s/\longs}. For \character{s/\longs}, even though not directly obvious due to the fusion of the two allographs, \character{\longs}'s foot is curved to the right (\inlinegraphics{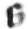}), a characteristic proper to NT. At the same time, \character{b}'s slightly dislocated lobe (\inlinegraphics{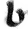}), also present in \character{h} (\inlinegraphics{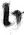}) and \character{p} (\inlinegraphics{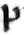}), characteristic of later examples of \textit{Textualis}~\cite{derolez2003palaeography}, is disrupting the typical circular lobe shape of ST. These dislocated lobes are clear in our Figure~\ref{fig:visual-comp}.

\paragraph{Later \textit{Textualis} examples.} 
Later specimens of both subtypes present notable differences from their earlier counterparts. \textbf{NT6}, copied in the last quarter of the 14\textsuperscript{th} c. is an example of later Northern \textit{Textualis}, and, while the graph shows that all prototypes are closer to the NT prototypes than the ST prototypes, it also clearly shows that they are very different from both, i.e., they are more on the top right of the graph. This document particularity lies in its strict angular forms, with diamond-shaped minim feet (\inlinegraphics{figures/south_north/sprites_text/NT6/36.png}\inlinegraphics{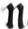}) as well as its exaggerated narrow shapes, with a total absence of round strokes for arcs (\inlinegraphics{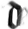}\inlinegraphics{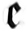}\inlinegraphics{figures/south_north/sprites_text/NT6/42.png}), typical of 14\textsuperscript{th}-15\textsuperscript{th} c. (esp. Northern) \textit{Textualis}. \textbf{ST7} is an example of late, 15\textsuperscript{th} c. Iberian \textit{Textualis}, discussed separately by Derolez due to its tendency towards more angular forms. Particularities of this type lie in the presence of hairlines and angular shapes (\inlinegraphics{figures/south_north/sprites_text/ST7/28.png}\inlinegraphics{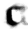}\inlinegraphics{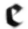}) alongside typical rounder ones (\inlinegraphics{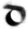}\inlinegraphics{figures/south_north/sprites_text/ST7/42.png}\inlinegraphics{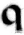}) with flat tops and feet~(\inlinegraphics{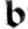}\inlinegraphics{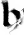}\inlinegraphics{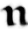}\inlinegraphics{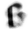}). However, these characteristics are only discernible while looking at the prototypes and their differences, and not directly in the graph, where the prototypes are closer to ST, and not particularly poorly reconstructed. This can be understood both by the fact that these particularities do not make the prototypes more similar to the NT ones, and by the fact that another Iberian \textit{Textualis} (ST4), from the same manuscript but from a different hand, was used in our reference set, and thus our ST prototype already models some characteristics of this \textit{Textualis} type. This highlights the impact of the prototype training data on our analysis, prototypes utilized as the basis for our comparison graph axes.

\section{Conclusion}
In this work, we introduced a deep learning-based methodology for interpretable script comparison and analysis. By applying it to the two subtypes of \textit{Textualis Formata} script type defined by A. Derolez —Northern and Southern \textit{Textualis}— we showed how such an approach can complement qualitative document analysis, by quantifying specific elements and summarizing information. We believe our approach contributes to bridging the gap between traditional and learning-based approaches to paleography.

\subsubsection{Aknowledgements} 
This study was supported by the CNRS through MITI and the 80|Prime program (CrEMe Caractérisation des écritures médiévales), and by the European Research Council (ERC project DISCOVER, number 101076028). We thank Ségolène Albouy, Raphaël Baena, Sonat Baltacı, Syrine Kalleli, and Elliot Vincent for valuable feedback.
%
%
%

\end{document}